\documentclass[conference]{IEEEtran}
\IEEEoverridecommandlockouts
\usepackage{amsmath,amssymb,amsfonts}
\usepackage{algorithmic}
\usepackage{graphicx}
\usepackage{textcomp}
\usepackage{xcolor}
\def\BibTeX{{\rm B\kern-.05em{\sc i\kern-.025em b}\kern-.08em
    T\kern-.1667em\lower.7ex\hbox{E}\kern-.125emX}}

\usepackage[ruled,vlined,linesnumbered]{algorithm2e}
\usepackage{mathtools}
\usepackage{float}

\usepackage{xr}
\usepackage[numbers]{natbib}
\usepackage{hhline}
\usepackage{hyperref}

\begin{document}

\title{IIFE: Interaction Information Based Automated Feature Engineering}

\author{\IEEEauthorblockN{Tom Overman}
\IEEEauthorblockA{\textit{Department of Engineering Sciences} \\\textit{
and Applied Mathematics} \\
\textit{Northwestern University}\\
Evanston, IL, USA \\
tomoverman2025@u.northwestern.edu}
\and
\IEEEauthorblockN{Diego Klabjan}
\IEEEauthorblockA{\textit{Department of Industrial Engineering}\\ \textit{and Management Sciences} \\
\textit{Northwestern University}\\
Evanston, IL, USA \\
d-klabjan@northwestern.edu}
\and
\IEEEauthorblockN{Jean Utke}
\IEEEauthorblockA{
\textit{Allstate Insurance Company}\\
jutke@allstate.com}
}

\maketitle

\begin{abstract}
  Automated feature engineering (AutoFE) is the process of automatically building and selecting new features that help improve downstream predictive performance. While traditional feature engineering requires significant domain expertise and time-consuming iterative testing, AutoFE strives to make feature engineering easy and accessible to all data science practitioners. We introduce a new AutoFE algorithm, IIFE, based on determining which feature pairs synergize well through an information-theoretic perspective called interaction information. We demonstrate the superior performance of IIFE over existing algorithms. We also show how interaction information can be used to improve existing AutoFE algorithms. Finally, we highlight several critical experimental setup issues in the existing AutoFE literature and their effects on performance.
\end{abstract}

\section{Introduction}
\label{intro}
Feature engineering is a technique used to craft new features that help downstream model performance \cite{feat_eng}. Until recently, feature engineering required significant domain expertise to create meaningful new features. Automated feature engineering attempts to automate the feature engineering process and allow general data science practitioners to benefit without requiring expert domain knowledge and time-consuming manual feature creation and testing. 

The feature engineering we consider in this work focuses on combining existing features together through the means of various bivariate operations, while also allowing transformations of single features, selecting features that are relevant, and removing features that are irrelevant. Feature engineering is often used to create nonlinear interactions between existing features that can help boost the performance of simple, linear models. As a very simple example, imagine building a predictive model to find the probability that a user buys a travel package to a Cancún hostel that is very popular among younger adults, and there are existing features such as user age and the frequency of web searches about Cancún vacations. A useful engineered feature for the model is $\frac{\text{freqCancunVacationSearch}}{\text{age}}$ which has a large value for the specific combination of high search frequency and young age and would penalize older vacation-searchers who are typically not interested in hostels. This allows a simple, linear model to capture this nonlinear interaction while still maintaining explainability. The goal of AutoFE is to find these, and even more complicated, engineered features in an automatic fashion without needing significant domain expertise and effort.

We propose a new AutoFE method based on interaction information. Interaction information is a way to calculate how well different feature pairs synergize in predicting a target \cite{ii_original}. While standard mutual information \cite{information_theory} can be used to calculate how much information is shared between a specific single feature and the target, interaction information expands upon this to quantify how much synergy is bound up in three variables (i.e. two features and the target). Therefore, pairs of features with high interaction information can be combined to create useful new features. Our algorithm, \textbf{I}nteraction \textbf{I}nformation Based Automated \textbf{F}eature \textbf{E}ngineering (IIFE), computes the interaction information between all possible pairs of features, then combines the highest synergy feature pairs using candidate uni- and bivariate functions. Then the highest scoring candidate engineered feature is added to the feature pool and the process is repeated, including the new engineered feature in the next iteration. In this way, our algorithm can create complex engineered features while only searching the feature pairs that synergize well, significantly reducing computation time.

Our contributions are as follows.
\begin{enumerate}
\item We develop an AutoFE algorithm, IIFE, focused around using interaction information to guide combining features. We create synthetic experiments to demonstrate that interaction information properly captures synergy between two features and the target. We then show that IIFE outperforms existing AutoFE methods on the relatively small public datasets used as benchmarks in the AutoFE literature and on a much larger private industry dataset. The code implementation is made open-source.
\item We demonstrate several experimental setup issues that affect most of the AutoFE literature, we quantify how large of an effect these issues have on reported scores, and we fix these issues in our algorithm comparison experiments.
\item We demonstrate that interaction information can be successfully incorporated into other expand-reduce AutoFE algorithms to accelerate these algorithms while maintaining similar or better downstream test scores.
\end{enumerate}

\section{Related Work}
The central problem in AutoFE is the exploding feature space as the order of transformations becomes larger. Finding a smart way of exploring this massive feature space is the main focus of most modern AutoFE algorithms. EAAFE uses a genetic algorithm approach to search the space \citep{eaafe}. Each chromosome corresponds to an original feature and the genes correspond to transformations applied to this feature. In the end, EAAFE doubles the size of the feature space. DIFER forms the feature transformations as strings and then uses deep learning techniques such as LSTMs to build a representation of the feature strings to predict validation performance \citep{difer}. OpenFE expands all possible nesting of transformations, up to a specified maximum order, but uses a LightGBM boosting technique to very quickly evaluate candidate features \citep{openfe}. Furthermore, OpenFE typically only allows transformations of order 2 to prevent the exploding feature problem. AutoFeat expands out all possible feature transformations, then uses a multi-step feature selection process to reduce the number of engineered features \citep{autofeat}. AutoFeat also typically requires the maximum order of the transformations to be relatively low to avoid the exploding number of features. There have also been reinforcement learning approaches to efficiently navigate the transformation graph \cite{rl}. NFS adapts techniques from neural architecture search \citep{nas} to effectively search for high-performing engineered features \cite{nfs}. These algorithms all use different techniques to search the tremendously large space of possible engineered features. We take a different approach by determining beforehand which feature pairs synergize well through interaction information and should be combined together.

\section{Algorithm Description}
Interaction Information is a generalization of mutual information to more than two variables and has been used in a variety of fields to represent the synergy bound up between multiple variables \cite{ii_bio}\cite{ii_astro}\cite{ii_original}. It was previously introduced to the data science literature to help find underlying interactions between features and to help select features to use directly in training predictive models \cite{interaction_information}. We take a new approach to use interaction information in feature engineering to determine which feature pairs to combine into new engineered features through bivariate functions; feature pairs that have high synergy in terms of interaction information are good candidates to combine together.

The interaction information of features $F_i,F_j\in \cal F$ and target $Y$ is computed as $\tau_{ij}=I(F_i,F_j,Y)=I(F_i,F_j|Y) - I(F_i,F_j)=H(F_i,F_j)+H(F_j,Y)+H(F_i,Y)-H(F_i)-H(F_j)-H(Y)-H(F_i,F_j,Y)$ where $I(F_i,F_j)$ is the standard mutual information of two variables and $H(F_i)$ is Shannon's entropy of $F_i$ \cite{information_theory}. Interaction information is symmetric in that $I(F_i,F_j,Y)=I(F_j,F_i,Y)=...=I(Y,F_i,F_j)$. Viewing the definition as $I(F_i,Y,F_j) = I(F_i,Y|F_j)-I(F_i,Y)$, it can be interpreted as finding the shared information between one feature and the target \textit{given} information about the other feature and subtracting the shared information between the feature and target without the influence of the other feature. In other words, it can be interpreted as how well two features synergize in predicting the target beyond just the shared information of the single variables and the target in isolation. 

\begin{figure*}[htbp]
\centerline{\includegraphics[width=.99\textwidth]{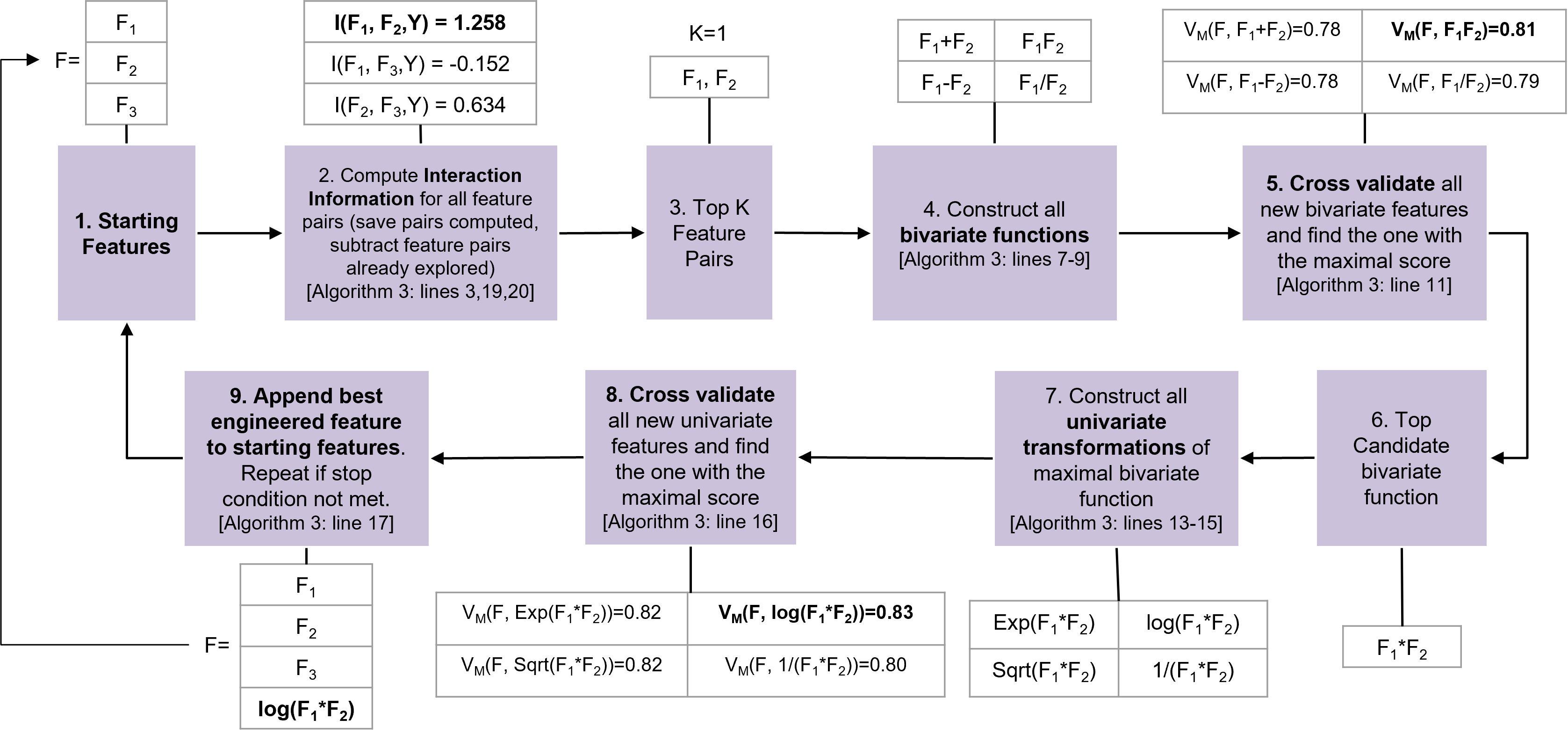}}
\caption{Flowchart of IIFE using a toy example of three starting features and few uni- and bivariate functions. In realistic settings there would be larger sets $\cal F,B$ and $\cal U$. The next iteration will include the newly engineered feature $\log(F_1 * F_2)$ in the pool of features, so increasingly complex features are engineered.}
\label{algorithm_flowchart}
\end{figure*}

Our proposed algorithm, IIFE, shown in Algorithm \ref{NewAlgo}, is an iterative process. In Fig. \ref{algorithm_flowchart} we depict a possible single iteration. First, interaction information values are computed for all pairs of original features. With the interaction information ${\cal O}(|{\cal F}|^2)$ compute complexity in the first iteration, $\cal F$ may have to be pre-filtered to  $\tilde{\cal F}\subseteq {\cal F}$ if $|{\cal F}|$ is too large. In subsequent iterations, the compute complexity of interaction information is ${\cal O}(|{\cal F}|)$. Then the algorithm explores feature pairs that have the highest interaction information. Combinations of the feature pairs using bivariate functions from a fixed set $\cal B$ are created and the downstream model $M$ performance is evaluated via cross validation $V_M$ in parallel. The best performing engineered feature is selected, and then univariate functions from a fixed set $\cal U$ of this new feature are computed and evaluated. This new feature is then added to the pool of features, and the process repeats. The next iteration, only the interaction information of the new feature with the existing features needs to be computed. In this iterative manner, engineered features can continuously build on previous features to generate new, more complex features. The combinations of features are determined solely by interaction information which characterizes how well two features synergize to predict the target.

\begin{algorithm}
\label{InteractionInformation}
\SetAlgoLined
Input: Set $\cal P$ of feature pairs $(F_i, F_j)$, labels/targets $Y$\\
${\cal I} \gets \emptyset$ \\
\For{$(F_i, F_j) \in \cal P$}{
    $\tau_{ij}=I(F_i,F_j,Y) = I(F_i,F_j|Y) - I(F_i, F_j)$ \\
    ${\cal I} \gets {\cal I} \cup \{(\tau_{ij},i,j)\}$
}
return $\cal I$
\caption{InteractionInformation (II)}
\end{algorithm}

\begin{algorithm}
\label{StopCondition}
\SetAlgoLined
Input: List of cross validation scores $S$, and stop patience $P$, current iteration count $c$\\
\eIf{$|S|\geq P$ and $\big(\underset{c - \frac{P}{2} < i \leq c}{mean}(S_i) - \!\!\!\!\!\underset{c - P < i \leq c-\frac{P}{2}}{mean}(S_i)\big)\leq 0$}{return True \tcp{Stop IIFE}}
{
return False \tcp{IIFE continue}}
\caption{StopCondition}
\end{algorithm}

\begin{algorithm}[htbp]
\label{NewAlgo}
\SetAlgoLined
Input: feature set $\tilde{\cal F}$, targets $Y$, feature pair count $K$ per iteration, patience $P$, model cross validation $V_M$\\
$S \gets [\,]$, $c=0$\\
${\cal I} \leftarrow \texttt{II}( \{ (F_i,F_j) | F_i, F_j\in \tilde{\cal F}, i\neq j \}, Y)$ \tcp{Algo\ref{InteractionInformation}}
\While{not \normalfont{\texttt{StopCondition}}($S,P,c$)}{
  $\mathcal{F}_{\cal B} \leftarrow \emptyset $\\
  \For{$K$ largest $\tau_{ij} \text{ from } (\tau_{ij},i,j) \in {\cal I}$}{
    \For{ $b \in {\cal B}$}{
        $\mathcal{F}_{\cal B} \gets \mathcal{F}_{\cal B} \cup \{b(F_i,F_j)\}$
    }
  }
$F'_{\cal B}=b'(F_{i'},F_{j'}) \in \underset{F\in {\cal F}_{\cal B}}{argmax}\;V_M(\tilde{\cal F}\cup \{F\},Y)$\\
${\cal F}_{\cal U} \leftarrow \{F'_{\cal B} \}$\\
\For{$u \in \cal U$}{
${\cal F}_{\cal U} \gets {\cal F}_{\cal U} \cup \{u(F'_{\cal B})\}$
}
$F'_{\cal U} \in \underset{F\in {\cal F}_{\cal U}}{argmax}\;V_M(\tilde{\cal F}\cup \{F\},Y)$\\
$\tilde{\cal F} \gets \tilde{\cal F} \cup \{F'_{\cal U}\}$.\\
$S \gets S \cup V_M(\tilde{\cal F},Y)$, $c=c+1$\\
${\cal I} \gets {\cal I} \setminus \{(\tau_{i'j'},i',j')\}$\\
${\cal I} \leftarrow {\cal I} \cup \{\texttt{II}(F'_{\cal U}\times (\tilde{\cal F} \setminus \{F'_{\cal U}\}), Y)\}$ \\
}
\caption{Interaction Information based Automated Feature Engineering (IIFE)}
\end{algorithm}

% \begin{algorithm}
% \label{StopCondition}
% \SetAlgoLined
% Input: Cross-validation scores of past $T$ iterations $s_i, s_{i-1}, ..., s_{i-(T-1)}$, stopping tolerance $\epsilon$ \\
% return mean($s_i, ..., s_{i-(T/2-1)}$) - mean(scores($s_{i-(T/2)},...,s_{i-(T-1)}$)) $\leq \epsilon$
% \caption{StopCondition}
% \end{algorithm}

\section{Experimental Results}

\subsection{Algorithm Comparisons}
\label{algo_comp}
We compare the performance of IIFE on public datasets with OpenFE, EAAFE, AutoFeat, and DIFER because they are recent state-of-the-art algorithms that provide open-source implementations. The public datasets vary in size and include classification and regression tasks. To demonstrate practical relevance, we test also on a  proprietary dataset that is several orders of magnitude larger than any dataset used in the AutoFE literature. We randomly chose the public datasets from the EAAFE, DIFER, and OpenFE papers to fairly compare to results reported in other work. Table \ref{dataset_summary} in the appendix shows the size and problem type of each dataset. For downstream models $M$ we choose linear models (logistic regression (LR) for classification and Lasso for regression), random forest (RF), and LightGBM (LGBM); this covers most of the commonly used models on tabular datasets.
\subsubsection{Experimental Setup}
The code for IIFE is available at the \textit{\textbf{completely anonymous}} GitHub link \url{https://github.com/2oppy67zj4ky/Appendix}. Specific details of algorithm settings and parameters for all of the AutoFE algorithms can be found in the Appendix. For each dataset and model a total of 25 runs are completed with different random seeds. There are five seeds for the train-test split (80\%-20\% split) to neutralize the effect of the specific train-test split on performance \cite{train_test_splits} and five seeds for the random components within the AutoFE algorithms, combined for a total of 25 runs. The only random component in IIFE is the random seed used to determine folds during cross-validation evaluations on the train set.

For each run, hyperparameter tuning is performed both before AutoFE and after AutoFE (with the new engineered features) which makes these experiments more realistic but also more computationally costly than experiments in prior work. During AutoFE model evaluations, we keep the hyperparameters constant at the value of tuning before AutoFE. Specifics of the hyperparameter tuning procedure are in the Appendix. The test score is found on the raw features as a baseline and is computed again after AutoFE. For classification, we use F1-micro as the test metric, and for regression, we use $1 - \text{(relative absolute error)}$ as the test metric, as used in \cite{eaafe}\cite{difer}. We record the elapsed time for each AutoFE algorithm as the time it takes to perform AutoFE and tune hyperparameters after AutoFE (each algorithm crafts different numbers of features). For the linear models we use one-hot encoding of the categorical variables for the public data and target encoding for the proprietary dataset. For linear models, we use min-max scaling of the features when performing evaluations.

\subsubsection{Public Data Results}

\begin{table*}[!htbp]
\caption{Summary of Performance of all AutoFE Methods Averaged over All Datasets. The best result across all algorithms is indicated in \textbf{bold.} Standard deviation is denoted with parenthesis (). The large standard deviation values indicate that while many datasets are improved substantially by AutoFE, some datasets are not impacted significantly.}
\begin{center}
\begin{tabular}{|c|r|r|r|r|r|r|}
\hline
\textbf{}&\multicolumn{6}{|c|}{\textbf{AutoFE Methods}} \\
\cline{2-7} 
\textbf{Performance Metric} & \textbf{\textit{Baseline}}& \textbf{\textit{IIFE} (ours)}& \textbf{\textit{OpenFE}} 
& \textbf{\textit{AutoFeat}} & \textbf{\textit{EAAFE}} & \textbf{\textit{DIFER}}
\\
\hline
Number of Top Ranks & 2 & \textbf{10} & 4 & 3 & 4 & 1 \\
\hline
Average Rank & 4.46 (1.86) & \textbf{2.50 (1.64)} & 3.38 (1.63) & 3.08 (1.50) & 3.13 (1.54) & 3.75 (1.33) \\
\hhline{|=|=|=|=|=|=|=|}
Average \% Change over Baseline & N/A & \textbf{26.88\% (90.6)} & 8.26\% (14.7) & 20.62\% (72.9) & 7.11\% (16.3) & 9.02\% (27.9) \\
\hline
Avg \% Change over Baseline (exc. OpenML586) & N/A & \textbf{8.74\% (21.2)} & 5.63\% (10.7) & 6.26\% (20.6) & 7.16\% (12.1) & 3.64\% (7.63) \\
\hline
Average \% Change over Baseline (Linear Only) & N/A & \textbf{72.42\% (153)} & 18.97\% (22.1) & 59.44\% (122) & 15.33\% (26.7) & 23.09\% (46.8) \\
\hline
Average \% Change over Baseline (RF Only) & N/A & \textbf{5.20\% (5.95)} & 2.69\% (3.59) & 1.59\% (2.70) & 3.32\% (3.97) & 1.79\% (3.24) \\
\hline
Average \% Change over Baseline (LGBM Only) & N/A & 3.02\% (4.54) & \textbf{3.11\% (3.89)} & 0.82\% (1.58) & 2.68\% (5.04) & 2.18\% (4.69) \\
\hline
% \multicolumn{7}{l}{$^{\mathrm{a}}$Sample of a Table footnote.}
\end{tabular}
\label{summary_performance}
\end{center}
\end{table*}

\begin{figure}[htbp]
\centerline{\includegraphics[width=.5\textwidth]{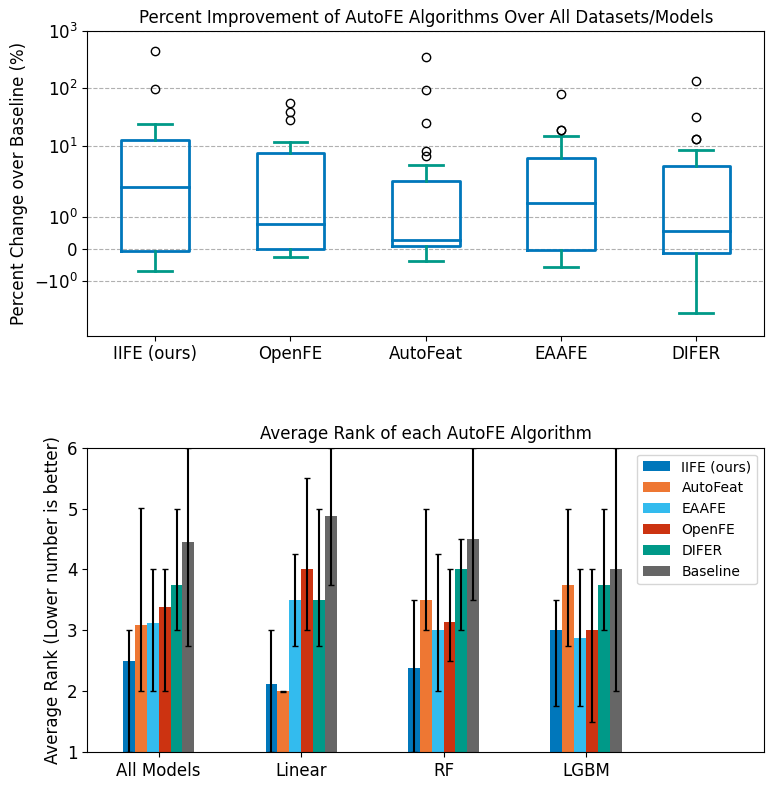}}
\caption{\textbf{Top:} Percent improvement of each algorithm over the baseline test score over all public datasets/models. The box represents the interquartile range with the central line being the median. The whiskers extend to the largest value within 1.5 times the interquartile range. The circles represent outliers. \textbf{Bottom:} Average rank for each algorithm over all of the public datasets and runs for all models, linear models, RF models, and LGBM models. Error bars show the 25th and 75th percentiles. The best performing algorithm would be rank 1, so lower average rank number is better.}
\label{algorithm_comparison}
\end{figure}

Table \ref{summary_performance} shows the key overall metrics for each AutoFE algorithm such as average percent change over the  baseline (original raw features) and average ranking of each algorithm. It is clear that our algorithm, IIFE, is the best performing method across all of the metrics except for the average percent change over baseline for LightGBM downstream models where OpenFE scores marginally better. Fig. \ref{algorithm_comparison} shows the percent improvements over the raw feature baseline and the average ranks of each of the algorithms for each downstream model (lower rank number is better). The 25th percentile is similar across all algorithms demonstrating that the bottom quarter of datasets/models do not improve from any of the AutoFE algorithms. However, IIFE (ours) has a substantially higher median and 75th percentile, demonstrating that IIFE's performance gains over other algorithms shown in Table \ref{summary_performance} are not just due to a few well-performing outliers. While the interquartile range is the largest for IIFE, the lower quartile is the same as the other algorithms and the variation extends solely in the positive direction. The runtimes of each of the AutoFE algorithms were similar, except for DIFER which took much longer on average. The bottom plot shows that across the different downstream models, IIFE is the top ranking model across most datasets. Table \ref{all_results} in the Appendix shows the full results and runtimes of our experiments on each individual dataset/model. These experiments demonstrate that IIFE is well-suited for a wide variety of datasets and downstream models and outperforms existing AutoFE methods on the majority of datasets.

\begin{figure*}[htbp]
\centerline{\includegraphics[width=.9\textwidth]{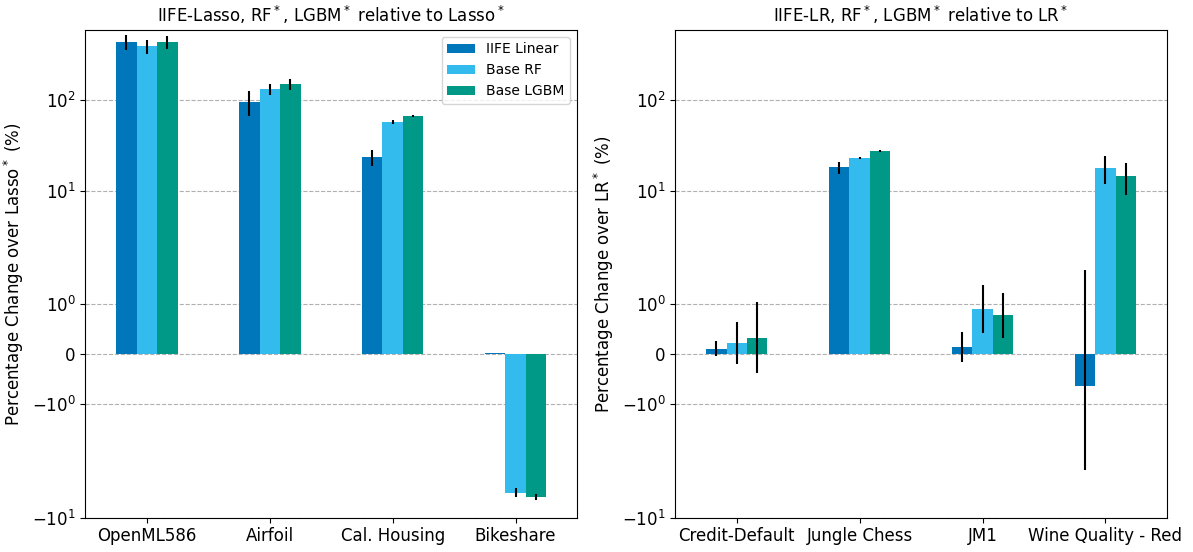}}
\caption{Plot demonstrating similar improvements between IIFE+Lasso, RF$^*$, and LGBM$^*$ relative to Lasso$^*$/LR$^*$. 
The $^*$ denotes models trained only on the original features. 
The errors bars are +/- the standard deviation across all 25 runs. This shows that on many datasets, engineering features with IIFE can bring linear models close to the performance of large, complicated nonlinear models such as random forest and LightGBM with large numbers of estimators and depth of trees.}
\label{linear_compare}
\end{figure*}

Tree-based models typically outperform linear models but we can show that using IIFE with linear models reduces the test score gap to RF and LGBM models. In practical settings, strict explainability requirements and regulations necessitate linear models and feature engineering enables linear model training on non-linear interactions. Therefore we consider this is a major contribution. We show the gap reductions in Fig. \ref{linear_compare} relative to RF$^*$/LGBM$^*$; the $^*$ models trained only on original features. The IIFE-improved LR models with the classification datasets reduce the average gap to RF$^*$ to 4.99\% and to 5.35\% to LGBM$^*$. Similarly, IIFE-improved Lasso models\footnote{The Bikeshare problem is an outlier in that it is perfectly linear.} with the regression datasets reduce the average gap to RF$^*$ to 4.94\% and to 10.15\% to LGBM$^*$. This demonstrates that linear models can be improved by IIFE to be comparable to complicated nonlinear models while still holding the advantages of simple linear models such as high explainability.

\begin{figure*}[htbp]
\centerline{\includegraphics[width=.9\textwidth]{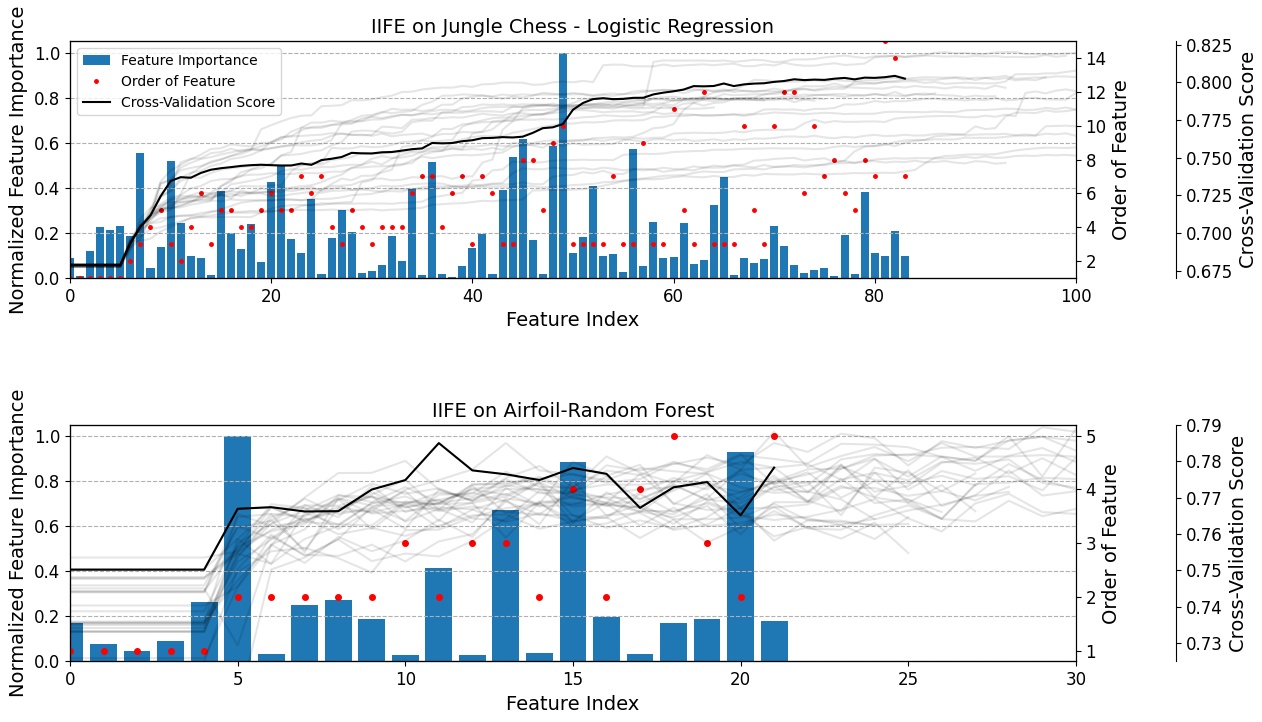}}
\caption{The blue bars show the feature importance, the red dots show the order of the feature, and the black line shows the cross-validation score after the feature was added. The faded lines in the background are the cross-validation curves for the other 24 runs with different random seeds. Some of the additional faded curves are truncated to focus the plot on the key iterations. \textbf{Top: } The plot shows a case where forming a large number of highly complex features helps performance. It also depicts the validation score growing as the number of features increases (the first several features of order 1 are the original features). The most important engineered feature is order 10 which is out of the practical complexity range for the majority of AutoFE algorithms.
\textbf{Bottom: } The plot shows a case where it is more beneficial to create fewer, low-order engineered features to boost validation scores. This is typically the case for more complicated models such as random forest and LightGBM which can already model complex non-linear behavior.
}
\label{feat_importance}
\end{figure*}

We now investigate possible reasons why IIFE outperforms existing AutoFE algorithms. The key difference between IIFE and other algorithms is that exploiting the interaction information component coupled with iterative feature construction favors the creation of specific high-order features while other algorithms cannot suitably reduce the feature search space that grows combinatorially with the feature order. The order of an engineered feature refers to the number of original features that constructs it. OpenFE limits engineered features to order 2, AutoFeat to order 3, EAAFE to order 6, and DIFER to order 5. Fig. \ref{feat_importance} shows two contrasting examples of IIFE on different datasets/models. For LR on the Jungle Chess dataset (the top figure), it is clear that adding relatively high order features is useful for improving performance as indicated by the most important feature being order 10 and the cross-validation scores steadily increasing as higher order features are added. For random forest regressor on the airfoil dataset (the bottom figure), IIFE only adds relatively low order features because there is no major benefit to adding highly complex engineered features in this case. 

This highlights a key strength of IIFE, the interaction information step and the iterative growth of engineered features allows the algorithm to build increasingly complex features if the model benefits from this, but can also build simple, low-order features if that is more suitable for the task at hand. Furthermore, if the user for interpretibility wishes to limit the maximal order of engineered features, then it is easy to adjust the algorithm to accommodate this by popping feature pairs in each iteration from the interaction information score list that have reached a threshold maximum order.

\subsubsection{Results on a large-scale proprietary data set}
To compare the scalability of AutoFE algorithms we tackle a regression problem on a large-scale proprietary dataset. This dataset has on the order of thousands of features, and hundreds of thousands of samples - much larger than the public datasets shown in the previous section and in the existing AutoFE literature. We solve a real-world problem with a downstream Lasso model. We tune the regularization constant before AutoFE and again after AutoFE when the newly engineered features are included.

Because of the size of the data in both samples and features, adjustments had to be made to the algorithms in order for them to run efficiently. For IIFE, each iteration we use a different random row subsampling when performing the evaluations, effectively reducing the sample size by a factor of around 20 to speed up the evaluations. More details of IIFE implementation on this dataset are provided in the Appendix. For both OpenFE and AutoFeat, we have to use the interaction information adjustments shown in Section \ref{sec:ii_improve}. Without these adjustments using interaction information to reduce the search space, standard OpenFE and AutoFeat cannot complete within several weeks of runtime. We rename the interaction information adjusted versions of these algorithms to OpenFE-II and AutoFeat-II. For all computations of interaction information (used in IIFE, AutoFeat-II, and OpenFE-II) we first select the 50 best features using RF impurity-based feature importance which form $\tilde{\cal F}$, and then we find the interaction information across all possible pairs of those 50 features. We do not compare with EAAFE and DIFER because there is no clear way to adjust the code using interaction information to reduce the search space, and these algorithms, in their original implementation, cannot run on this large dataset.

Each algorithm is tested with 5 runs where each run uses a different random seed for the algorithm and the train-test split. Table \ref{private_results} shows the performance of the various AutoFE algorithms on this large-scale dataset, averaged over the 5 runs. It is clear that IIFE is the best performing method on this large-scale dataset.

\begin{table}[htbp]
\caption{Large-Scale Private Data Results. Quantities in parenthesis () are standard deviations.}
\begin{center}
\begin{tabular}{|c|r|r|}
\hline
\textbf{Algorithm} & \textbf{\textit{\% Change Over Baseline}}& \textbf{\textit{Runtime (hour)}} 
\\
\hline
IIFE (ours) & \textbf{6.2137\%} & 5.33 (0.28) \\
\hline
OpenFE-II  & 1.1649\% &  4.17 (0.22) \\
\hline
AutoFeat-II  & 4.2267\% & 5.75 (0.95) \\
\hline
\end{tabular}
\label{private_results}
\end{center}
\end{table}

\subsection{Experimental Verification of Interaction Information}
\begin{figure}[htbp]
\centerline{\includegraphics[width=.5\textwidth]{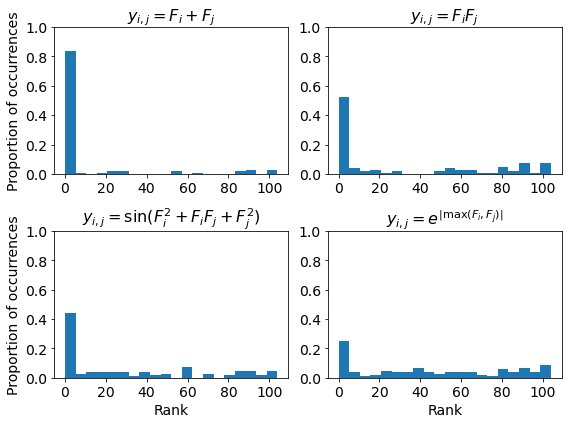}}
\caption{Histograms of rank of the true feature pair when computing the interaction information of the feature pair with the synthetic target built from the feature pair. This is repeated across all possible feature pairs. We expect the rank to be zero or a low number and most of the density of the histogram to be on the left side. The synthetic target is constructed for $F_i +F_j$, $F_i F_j$, $\sin(F_i^2 + F_ix_j + F_j^2)$, and $\exp(|\max(F_i,F_j)|)$ which are increasingly complex, nonlinear, and in the final example, non-smooth.}
\label{ii_experiments}
\end{figure}

In order to verify that interaction information can be used to determine the synergy of two features in predicting a target, we develop a synthetic data experiment. We create synthetic targets $y_{ij}=f(F_i,F_j)$ as functions $f$ of two of the input features $F_i, F_j$ from the credit-approval dataset. We then find the interaction information of all feature pairs $(F_l,F_k)$ with the synthetic target $y_{ij}$. After this we find the rank of the interaction information $I(F_i,F_j,y_{ij})$ and store this rank. We expect the ranks to be a low number (the best rank is rank 0) indicating the highest interaction information). We then repeat this entire process for all 105 feature pairs in that dataset. In Fig.\ref{ii_experiments} we show the histograms for exemplary $f$. A histogram with most mass on the left side of the plot indicates that interaction information works as expected. Fig. \ref{ii_experiments} validates that interaction information is a good proxy for synergy between two features even for complex, nonlinear relationships.

\subsection{Issues in AutoFE Literature}
We have identified several issues in the AutoFE literature.
\begin{itemize}
    \item The vast majority of AutoFE papers do not use a hold-out test set and instead report final metrics as cross-validation scores on the full dataset. It has been shown that overfitting can still occur when using cross-validation especially when optimizing the engineered features and reporting final performance scores on the same cross-validation scheme \citep{cv_overfit}.
    \item OpenFE is the only paper to use a hold-out test set but operates in the ``transductive learning" scheme where certain aggregation and scaling operations are performed on the combined train and test sets which clearly reveals information about the test samples into the AutoFE process. This scheme overestimates performance of models deployed in data-limited settings when aggregates for new data no longer perfectly match the aggregates established during training in the transductive scheme.
    \item Most papers do not perform hyperparameter tuning before and after the AutoFE process. This does not reflect a typical process.
\end{itemize}

\subsubsection{Cross-validation scores as performance metric}
We show that cross-validation (CV) scores on the full set of data is not a good measure of AutoFE performance. Instead, the more realistic setting of using a hold-out test set is necessary. For the CV measure  we follow the procedure in \cite{eaafe} where the CV scores are used as measures of fitness for the genetic algorithm and these same CV scores are used as the final metric for reporting how well EAAFE performs. We contrast this with the alternative and better aligned with practice approach which is to use a hold-out test set and find the test score as the final metric. For each dataset, we conduct 25 runs with different random seeds and find averages. Averaged over all of the datasets for the linear downstream model, the percent increase with the CV scheme is 21.0348\%, but the hold-out test score percent improvement is only 15.3270\%. This demonstrates that the results in \cite{eaafe}\cite{difer}, and other works that report only CV scores should be taken with a grain of salt.

\subsubsection{OpenFE transductive setting}
\label{transductive}
OpenFE operates in the transductive learning setting, which allows the algorithm to use test samples (but not the test labels) in constructing new features. The groupbythen* operations all utilize information of test samples in transforming the test and train features. However, in practice, to get more robust estimates of model performance, we typically operate in the inductive setting and do not have access to the test samples. To fix these groupbythen* operations, the groups and corresponding aggregation values are determined solely on the training data and then applied to the test data.

We adjust OpenFE to operate in the inductive setting rather than the transductive setting. For each dataset, we conduct 25 runs with different random seeds and find the test scores on the original version of OpenFE and the adjusted version of OpenFE. For most of the datasets, there is not a significant difference in scores between the transductive and the inductive version; however, for the airfoil dataset, the original, transductive version of OpenFE has a test score 14.01\% higher than the adjusted, inductive version of OpenFE. For the jungle chess dataset, the original, transductive version of OpenFE has a test score 1.95\% higher than the adjusted, inductive version of OpenFE. This suggests that further comparisons with OpenFE and other algorithms that utilize functions that reveal information about the test samples into the training data, such as scaling or aggregation functions on combined train and test data, need to be redone to ensure a fair comparison.

\subsection{Improving other algorithms with interaction information}
\label{sec:ii_improve}For algorithms that use an expand-reduce framework where the possible feature combinations are computed and then feature selection is applied to reduce the number of engineered features, interaction information can be used to significantly reduce the number of feature combinations that are explored. Interaction information is used to determine the highest synergizing feature pairs, and only functions of those high interaction information feature pairs should be expanded in the algorithm. Since AutoFeat and OpenFE both follow the expand-reduce framework, we explore using interaction information to accelerate these algorithms. 

\begin{figure}[htbp]
\centerline{\includegraphics[width=.5\textwidth]{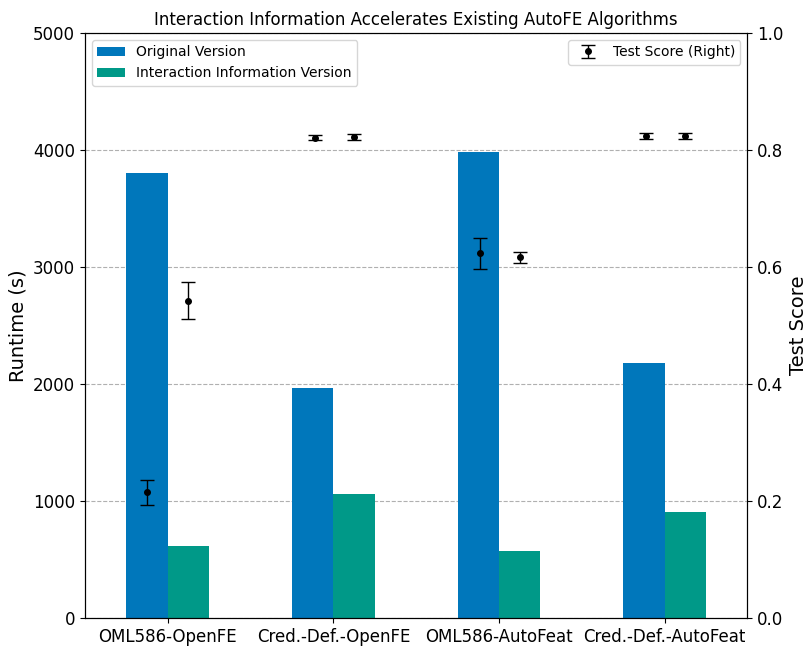}}
\caption{Runtimes and test scores for the original version of AutoFE and OpenFE and the interaction information (II) accelerated versions across two datasets. The results show similar or better test score performance with much shorter runtimes due to the reduction in search space that II provides. The runtimes include the additional cost of computing II for the accelerated version.}
\label{ii_improve}
\end{figure}

Fig. \ref{ii_improve} illustrates the runtime decrease that the accelerated versions  achieve along with the black error bars for test scores remaining the same or even improving for the accelerated version. For both datasets and algorithms, we use interaction information to reduce the set of feature pairs by about a factor of five.  For each dataset, we conduct 25 runs with different random seeds and show averages. The plot shows that the test scores of the accelerated version are comparable or better than the original version with significantly shorter runtimes. These results show that reducing the search space with interaction information can help make AutoFE algorithms faster without a degradation in downstream model performance.

\subsection{Combining AutoFE algorithms} Each AutoFE algorithm generates different engineered features, so we can combine the engineered features of different AutoFE methods to get better performance than either of the methods by themselves. We combine the engineered features of the two top performing AutoFE algorithms for each downstream model. On average, IIFE+AutoFeat for LR is 1.51\% higher, IIFE+EAAFE for RF is 0.31\% higher, and IIFE+OpenFE for LGBM is 0.20\% higher than the second best scoring method. This shows that combining AutoFE methods together helps the most with linear models.

\section{Conclusion}
We present IIFE, an iterative, open-source AutoFE algorithm that uses interaction information to determine which pairs of features to combine, significantly reducing the search space and allowing complex and high-performance engineered features to be found in short amounts of time. We demonstrate empirically that interaction information can identify pairs of features that synergize well with the target. We then show the superiority of IIFE over other AutoFE algorithms on a multitude of public datasets and a large-scale industry dataset. The percent improvement for IIFE over the original, raw features averaged over all datasets/downstream models is 26.88\%. We show that IIFE is adaptable and can find more complex, higher-order features than existing AutoFE methods, while it also constructs simple, low-order features if the downstream model benefits more from these. We further demonstrate that interaction information can significantly accelerate existing expand-reduce AutoFE algorithms. We also note several experimental setup issues across the AutoFE literature, how they inflate reported improvements in score, and how we fix them in our experiments. 

\bibliographystyle{plain}
\bibliography{refs} % Entries are in the refs.bib file
\nocite{*}

\section{Appendix}
\subsection{Full Experimental Results}
We show full results in Table \ref{all_results} and runtimes in Table \ref{runtimes}.

\begin{table}
\caption{Summary of Datasets}
\begin{center}
\begin{tabular}{|c|r|r|c|}
\hline
\textbf{Dataset} & \textbf{\textit{\# Samples}}& \textbf{\textit{\# Features}}& \textbf{\textit{Type}} 
\\
\hline
Airfoil \cite{airfoil} & 1,503 & 5 & Regression  \\
\hline
Credit Default \cite{credit_default} & 30,000 & 23 & Classification \\
\hline
Bikeshare \cite{bikeshare} & 17,389 & 13 &  Regression \\
\hline
Wine Quality - Red \cite{winequality-red} & 999 & 12 & Classification \\
\hline
California Housing \cite{california_housing} & 20,640 & 8 & Regression \\
\hline
OpenML 586 \cite{fri} & 1,000 & 25 & Regression \\
\hline
JM1 \cite{jm1} & 10,885 & 22 &  Classification \\
\hline
Jungle Chess \cite{jungle_chess} & 44,819 & 6 & Classification \\
\hline
Large-Scale Private Data  & $\approx 100,000$s & $\approx 1,000$s & Regression \\
\hline
% \multicolumn{7}{l}{$^{\mathrm{a}}$Sample of a Table footnote.}
\end{tabular}
\label{dataset_summary}
\end{center}
\end{table}

\begin{table*}
\caption{Performance of Different AutoFE Algorithms across different datasets and downstream models. Quantity in parenthesis () is standard deviation. WineQuality-Red is N/A for OpenFE because the OpenFE implementation returned an error pertaining to the LGBM validation early stopping failing.}
\begin{center}
\begin{tabular}{|c|c|c|c|c|c|c|}
\hline
\textbf{}&\multicolumn{6}{|c|}{\textbf{AutoFE Methods}} \\
\cline{2-7} 
\textbf{Dataset/Model} & \textbf{\textit{Baseline}}& \textbf{\textit{IIFE (Ours)}}& \textbf{\textit{OpenFE}} 
& \textbf{\textit{AutoFeat}} & \textbf{\textit{EAAFE}} & \textbf{\textit{DIFER}}
\\
\hline
Cal. Housing (Lasso)& 0.4111 (0.0053)& 0.5076 (0.0178) & \textbf{0.5260 (0.0057)} & 0.5137 (0.0069) & 0.4895 (0.0149) & 0.4651 (0.0076) \\
Cal. Housing (RFR)& 0.6476 (0.0053)& \textbf{0.7001 (0.0109)} & 0.6668 (0.0066) & 0.6584 (0.0055) & 0.6978 (0.0071) & 0.6596 (0.0315) \\
Cal. Housing (LGBM-reg)& 0.6861 (0.0048)& 0.6844 (0.0054) & 0.6958 (0.0044) & 0.6893 (0.0051) & \textbf{0.6967 (0.0051)} & 0.6859 (0.0195) \\
\hline
OpenML 586 (Lasso)& 0.1383 (0.0188)& \textbf{0.7494 (0.0520)} & 0.2150 (0.0212) & 0.6235 (0.0264) & 0.1645 (0.0643) & 0.3261 (0.1230) \\
OpenML 586 (RFR)& 0.6846 (0.0050)& \textbf{0.7803 (0.0165)} & 0.7522 (0.0062) & 0.7397 (0.0084) & 0.6939 (0.0244) & 0.6992 (0.0372) \\
OpenML 586 (LGBM-reg)& 0.7483 (0.0156)& 0.7907 (0.0141) & \textbf{0.7988 (0.0044)} & 0.7826 (0.0107) & 0.7483 (0.0189) & 0.7651 (0.0200) \\
\hline
JM1 (LR)& 0.8137 (0.0030)& 0.8149 (0.0046) & 0.8141 (0.0057) & \textbf{0.8161 (0.0020)} & 0.8146 (0.0033) & 0.8153 (0.0042) \\
JM1 (RF)& 0.8210 (0.0025)& 0.8209 (0.0039) & 0.8196 (0.0070) & \textbf{0.8227 (0.0012)} & 0.8195 (0.0034) & 0.8209 (0.0038) \\
JM1 (LGBM-class)& 0.8200 (0.0021)& 0.8175 (0.0022) & \textbf{0.8220 (0.0028)} & 0.8169 (0.0022) & 0.8184 (0.0029) & 0.8186 (0.0023) \\
\hline
Jungle Chess (LR)& 0.6765 (0.0028)& \textbf{0.7988 (0.0178)} & 0.7543 (0.0050) & 0.7207 (0.0042) & 0.7135 (0.0109) & 0.7080 (0.0064) \\
Jungle Chess (RF)& 0.8307 (0.0027)& \textbf{0.9396 (0.0335)} & 0.8391 (0.0045) & 0.8328 (0.0035) & 0.9148 (0.0367) & 0.8999 (0.0335) \\
Jungle Chess (LGBM-class)& 0.8617 (0.0039)& 0.9701 (0.0244) & 0.9481 (0.0113) & 0.8639 (0.0029) & \textbf{0.9873 (0.0059)} & 0.9737 (0.0114) \\
\hline
Airfoil (Lasso)& 0.3294 (0.0258)& \textbf{0.6468 (0.0819)} & 0.4547 (0.0248) & 0.6349 (0.0244) & 0.5871 (0.0529) & 0.4318 (0.0344) \\
Airfoil (RFR)& 0.7678 (0.0119)& 0.7834 (0.0139) & 0.7723 (0.0070) & 0.7753 (0.0082) & \textbf{0.7923 (0.0105)} & 0.7543 (0.0236) \\
Airfoil (LGBM-reg)& 0.8233 (0.0139)& \textbf{0.8386 (0.0152)} & 0.8260 (0.0132) & 0.8257 (0.0149) & 0.8372 (0.0155) & 0.8311 (0.0101) \\
\hline
Credit-Default (LR)& 0.8228 (0.0052)& \textbf{0.8237 (0.0061)} & 0.8209 (0.0043) & 0.8235 (0.0053) & 0.8218 (0.0058) & 0.8229 (0.0055) \\
Credit-Default (RF)& 0.8247 (0.0056)& 0.8242 (0.0050) & \textbf{0.8254 (0.0065)} & 0.8248 (0.0055) & 0.8250 (0.0056) & 0.8212 (0.0062) \\
Credit-Default (LGBM-class)& \textbf{0.8255 (0.0057)}& 0.8252 (0.0057) & 0.8239 (0.0046) & 0.8254 (0.0057) & 0.8250 (0.0056) & 0.8239 (0.0062) \\
\hline
Bikeshare (Lasso)& 0.9999 (0.0000)& 0.9999 (0.0000) & 0.9998 (0.0000) & 0.9999 (0.0000) & 0.9999 (0.0000) & \textbf{1.0000 (0.0000)} \\
Bikeshare (RFR)& 0.9471 (0.0057)& \textbf{0.9955 (0.0011)} & 0.9898 (0.0005) & 0.9628 (0.0064) & 0.9931 (0.0010) & 0.9878 (0.0088) \\
Bikeshare (LGBM-reg)& 0.9407 (0.0048)& \textbf{0.9864 (0.0042)} & 0.9708 (0.0028) & 0.9511 (0.0044) & 0.9795 (0.0073) & 0.9751 (0.0099) \\
\hline
Wine Quality-Red (LR)& 0.6050 (0.0200)& 0.6011 (0.0176) & N/A & 0.6056 (0.0237) & \textbf{0.6105 (0.0188)} & 0.6040 (0.0215) \\
Wine Quality-Red (RF)& \textbf{0.7138 (0.0294)}& 0.7089 (0.0187) & N/A & 0.7131 (0.0201) & 0.7099 (0.0204) & 0.7132 (0.0209) \\
Wine Quality-Red (LGBM-class)& 0.6938 (0.0253)& 0.6925 (0.0179) & N/A & \textbf{0.6956 (0.0215)} & 0.6919 (0.0186) & 0.6799 (0.0226) \\
\hline
% \multicolumn{7}{l}{$^{\mathrm{a}}$Sample of a Table footnote.}
\end{tabular}
\label{all_results}
\end{center}
\end{table*}

\begin{table*}
\caption{Runtimes in Hours (hr) of Different AutoFE Algorithms averaged across all datasets.}
\begin{center}
\begin{tabular}{|c|r|r|r|r|r|r|r|r|r|r|}
\hline
\textbf{}&\multicolumn{2}{|c|}{\textbf{IIFE (Ours)}}&\multicolumn{2}{|c|}{\textbf{OpenFE}} &\multicolumn{2}{|c|}{\textbf{AutoFeat}}&\multicolumn{2}{|c|}{\textbf{EAAFE}}&\multicolumn{2}{|c|}{\textbf{DIFER}}\\
\cline{2-11} 
\textbf{Model} & \textbf{\textit{Time (hr)}}& \textbf{\textit{Std Dev}} 
& \textbf{\textit{Time (hr)}}& \textbf{\textit{Std Dev}} & \textbf{\textit{Time (hr)}}& \textbf{\textit{Std Dev}} & \textbf{\textit{Time (hr)}}& \textbf{\textit{Std Dev}}& \textbf{\textit{Time (hr)}}& \textbf{\textit{Std Dev}}
\\
\hline
TOTAL Average& 1.58 & 1.51 & 1.45 & 2.41 & 1.53 & 1.59 & 1.31 & 1.46 & 4.55 & 3.16  \\
\hline
Average (Linear)& 1.52 & 1.73 & 0.67 & 2.41 & 0.61 & 0.33 & 0.33 & 0.07 & 1.71 & 0.36  \\
\hline
Average (RF)& 1.05 & 1.10 & 0.92 & 3.91 & 1.34 & 0.48 & 2.00 & 2.75 & 5.46 & 4.01 \\
\hline
Average (LGBM-class)& 2.29 & 2.15 & 2.74 & 2.60 & 2.62 & 3.07 & 1.35 & 0.81 & 7.07 & 4.75  \\
\hline
% \multicolumn{7}{l}{$^{\mathrm{a}}$Sample of a Table footnote.}
\end{tabular}
\label{runtimes}
\end{center}
\end{table*}

\subsection{Implementation Details}
Section \ref{algo_comp} has a link to the source code.

\textbf{Hardware: } All public datasets were run on Intel Xeon Silver 4108 CPUs @ 1.80 GHz. For algorithms using Ray, 8 cores on a single CPU were utilized to parallelize operations. For DIFER, the only implementation using a GPU, a single NVIDIA GeForce GTX 1080 Ti was used.

\textbf{Hyperparameter Tuning: } We tune the parameters before and after AutoFE. We use the same range of hyperparameters for all algorithms. We employ the Scikit-learn implementation of Random Search to randomly draw samples and 5-fold cross-validation, 100 iterations, and $\texttt{random\_state}=0$. For LR we tune $C$, for Lasso we tune $\alpha$, and both draw random values from loguniform$(0.00001,100)$. For both RF regressor and classifier, we tune \texttt{max\_depth} from randint$(1,250)$, \texttt{n\_estimators} from randint$(5,250)$, \texttt{max\_features} from uniform$(0.01,0.99)$, and \texttt{max\_samples} from uniform$(0.1,0.9)$. We tune LGBM classifier and regressor parameters as \texttt{n\_estimators} from randint$(10,1000)$, \texttt{learning\_rate} from loguniform$(.001,1)$, \texttt{subsample} and \texttt{colsample\_bytree} from uniform$(.1,.9)$, \texttt{reg\_lambda} from loguniform$(.001,100)$, and \texttt{num\_leaves} from randint$(8,64)$. These ranges are typical in prior work and are even larger than those presented in \cite{openfe}.

\textbf{IIFE: }  For public data, we do not filter the features before computing interaction information (II). Since II is just the difference of a conditional mutual information (CMI) and mutual information (MI), the II is computed using the CMI estimator provided in \cite{cmi} which provides a nearest neighbors approach for computing CMI with continuous and discrete variables and with a random subset of 3,000 samples. Evaluations are performed in parallel using Ray. For the public data we use $K=3$ and a stop patience $P$ of 20. For LR/Lasso, we speed up evaluations by using \texttt{max\_iter=100}. For cal. housing, openml586, and jm1 we set $P=40$. For Lasso, we remove the $\div$ and $1/x$ operators to avoid divisions by 0. For RF models for cal. housing, jm1, and jungle chess, we decrease \texttt{n\_estimators}=50 and \texttt{max\_samples}=0.25 due to the large size of these datasets. For lines 11 and 16 of Algorithm \ref{NewAlgo}, we use numpy.argmax which takes the first index of possibly multiple occurrences of the maximum value where the order is based on the array of input values. We allow the following bivariate functions:
\begin{enumerate}
    \item \textbf{Num-Num}: +, -, *, \, min, max, $x_1/(abs(x_2)+1)$, \%, and all reverse operations for non-commutative operations
    \item \textbf{Num-Categorical and Categorical-Num}: groupbythen- \{min,max,mean,median,std\} and all num-num operations if we treat the categorical variable as ordinal. The groupbythen* operations group the data by the categorical value, then find the aggregation operation of the numerical column for each group. For categories that are in test but not in the training data and numerical values are needed for each category, the value of the maximum absolute value in the training data plus $10\%\cdot i$ of this value is assigned for the $i$-th such category. For simplicity in our implementation, we assign the value of 0.
    \item \textbf{Categorical-Categorical}: Same as cat-num and include num-num operations if we treat the columns as ordinal.
\end{enumerate}
For univariate functions of argument $x$ we include $x^2, |x|, \sqrt{|x|}, \frac{1}{1+\exp(-x)}, 1/x$.

\textbf{OpenFE: } We fix the implementation of OpenFE as mentioned in Section \ref{transductive}. Due to the very large number of features created by OpenFE, it becomes prohibitively expensive to re-tune the parameters with the expanded feature space. So, we use the hyperparameters found before the OpenFE process for the RF and LGBM models for the larger datasets credit-default, openml586, jm1, jungle chess, and cal. housing.

\textbf{AutoFeat: } We use all default AutoFeat parameters except we change $\texttt{feateng\_steps}=1$ only for the large datasets credit-default, jm1, and jungle chess which took too long with the default parameters.

\textbf{EAAFE: } We used the Ray version, \texttt{n\_generations}$=5,000$, \texttt{c\_orders, n\_orders}$=5$, \texttt{pop\_size}$=16$, \texttt{cross\_rate}$=0.4$, and \texttt{mutate\_rate}$=0.1$. For LGBM runs and cal. housing, jungle chess, openml 586, credit-default, and jm1 we set \texttt{n\_generations}$=500$ and \texttt{pop\_size}$=8$ to bring the runtimes to a reasonable length.

\textbf{DIFER: }
We use default settings and early stopping as mentioned in \cite{difer}. 
For the large datasets OpenML586, JM1, Jungle Chess, and Cal. housing for RF and LGBM models we set the parameters as \texttt{top\_feat,new\_feat,random\_set\_size}=16 and \texttt{patience}=10 to bring the runtimes to a reasonable length. 

\textbf{Large-scale Company-Owned Data: }
We use target-encoding with a train cross-fitting scheme with 5 folds to encode the categorical variables because the feature dimension would increase too large using one hot encoding. Due to the large size dataset, we use random subsampling (by about a factor of 20 with a different seed each iteration) on steps 5 and 8 on Fig. \ref{algorithm_flowchart}, the evaluations of the candidate uni- and bivariate functions, to perform the evaluations more quickly.

\subsection{On the Lack of Performance Improvement on Wine Quality - Red (WQR)}
On the WQR dataset (target is the rating 0-10), we observe several AutoFE methods end with worse test scores than baseline. We explore this by reframing the problem as a regression problem and using IIFE. The percent change over baseline (\%OB) for the original classification problem with LR was -0.6446\%. The \%OB for the regression model with regression metric is 4.4318\%. The \%OB for the regression model with the decision thresholding and classification metric is 2.2312\%. This demonstrates that the poor performance on WQR may be due to a poor model training approach rather than a deficit in the AutoFE methods.

\end{document}